\documentclass{ecai}
\usepackage{times}
\usepackage{graphicx}
\usepackage{latexsym}

\usepackage{array}
\usepackage{soul}
\usepackage{color}
\usepackage[utf8]{inputenc}
\usepackage{amsmath}
\newcommand{\figref}[1]{Figure \ref{#1}}

\newcommand{\secref}[1]{Section \ref{#1}}

\renewcommand\appendix{\setcounter{secnumdepth}{-2}}

\newcommand{\tabincell}[2]{\begin{tabular}{@{}#1@{}}#2\end{tabular}}
\usepackage{diagbox}
\usepackage{multirow}
\usepackage{amsthm}
\usepackage{mathtools}
\usepackage{hhline}
\usepackage{booktabs}
\usepackage{makecell}
\usepackage{amssymb}
\usepackage{url}
\usepackage{CJKutf8}

\begin{document}

\title{Matching Questions and Answers in Dialogues from Online Forums}

\author{Qi Jia~$^1$ \and Mengxue Zhang~$^1$  \and 
Shengyao Zhang \institute{Shanghai Jiao Tong University, China} \and 
Kenny Q. Zhu \institute{Shanghai Jiao Tong University, China, email: 
kzhu@cs.sjtu.edu.cn (Contact author)}}

\maketitle
\bibliographystyle{ecai}

\begin{abstract}
 Matching question-answer relations between two turns in conversations is not only 
the first step in analyzing dialogue structures, 
but also valuable for training dialogue systems. 
This paper presents a QA matching model considering both 
distance information and dialogue history by two simultaneous attention 
mechanisms called \textit{mutual attention}. 
Given scores computed by the trained model between each non-question 
turn with its candidate questions, a greedy matching strategy 
is used for final predictions. 
Because existing dialogue datasets such as the Ubuntu dataset are not 
suitable for the QA matching task, we further create a dataset with 
1,000 labeled dialogues and 
demonstrate that our proposed model outperforms the state-of-the-art and 
other strong baselines, particularly for
matching long-distance QA pairs.
\end{abstract}

\section{Introduction}
\label{sec:intro}
Question motivated dialogues are very common in daily life and 
they are rich sources for question-answer (QA) pairs. 
For example, in an online forum for health consultations, 
both the doctor and the patient tend to ask and answer questions to 
narrow down the information gap and reach the final diagnosis or 
recommendations. Matching QA pairs can help track the final answers from the doctor
to the original patient question and is valuable for the medical
domain.

QA matching is an important part of analyzing discourse structures 
for dialogue comprehension. Asher et. al~\cite{asher2016discourse} 
shows that in online dialogues where participants are prompted to 
communicate with others to achieve their goals, 24.1\% of the relations 
between elementary discourse units are QA pairs. 
Questions and answers are the main components of 
dialogue acts~\cite{stolcke2000dialogue}, providing key features for 
dialogue summarization and decision detection~\cite{fernandez2008modelling}. 
Besides, figuring out the QA relations between these utterances can 
provide question answering 
models~\cite{ji2014information,vinyals2015neural,cui2017superagent} with 
high-quality QA pairs and contribute to the exploration of proactive questioning \cite{yan2017building}.

\begin{figure}[t]
	\centering
	\includegraphics[scale=0.29]{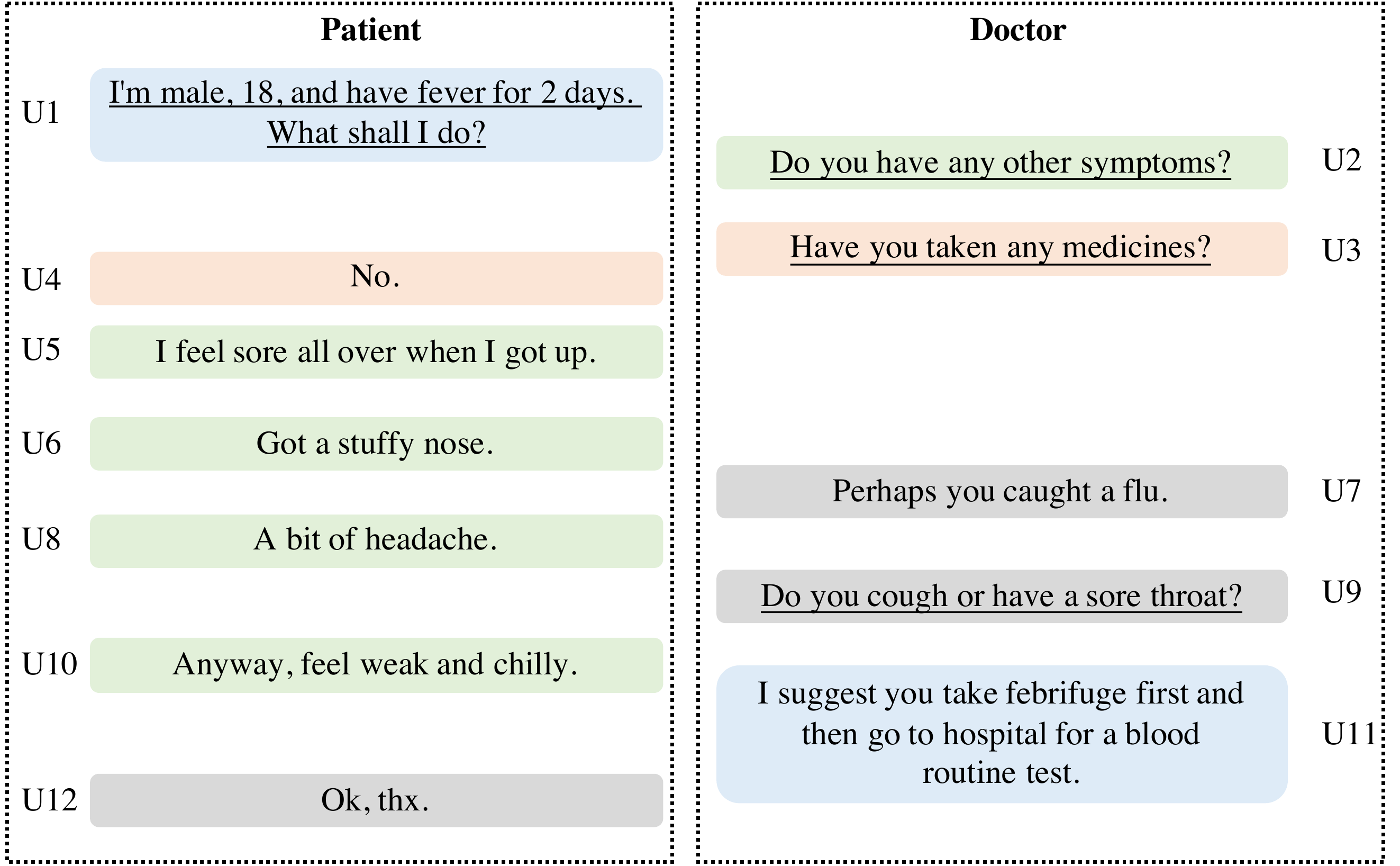}
	\caption{Questions and answers matching in dialogues from an online health forum. The identified pairs are painted in the same color and questions are underlined.}
	\label{fig:sample}
\end{figure}

However, many challenges exist. While it is relatively easy to distinguish
between questions and non-questions~\footnote{This can be done with a simple 
neural-based classifier with high accuracy.},
the non-questions may contain not only valid answers, but also chit-chats and
other informative statements. It is also a common phenomenon that a long and complete answer is broken up into several turns such as \{U5,U6,U8,U10\} in \figref{fig:sample}. Due to factors such as the network delay and different typing speed, the dialogue sequences are always mix-matched. Moreover,``personalized" orthography, ellipses, abbreviations, and missing punctuations are all difficulties for QA matching.

In this work, we focus on the task of matching questions and answers in two-party 
multi-turn dialogues. We found that the distance between the question and its answers 
is not only caused by the mix-matching and fragmentation mentioned above, 
but also by the very nature of the question. 
Some questions can be answered directly based on personal knowledge, 
such as U3, while others can not. For instance, when a patient asks questions 
such as ``what's wrong with me'' or ``what should I do'' just like U1, the doctor often has to ask follow-up questions \{U2,U3\} to seek additional information in order to give the final diagnosis or recommendation (U11). This often has to be done with several rounds of communication. 
We call this kind of QA pairs {\em incremental QA}. 
Such QA pairs often form the main idea of a dialogues or sub-dialogue, 
critical for dialogue comprehension. The answers are inherently far from the question (distance $\geq$ 3 \footnote{There is at least one follow-up question and one corresponding answer between the question and answer of an incremental QA pair. So the distance for such QA pairs is larger or equal to 3.}), aggravating the difficulty of matching such pairs. 

Roughly, we can categorize QA pairs according to the distance between them. 
When distance $\leq$ 3, we call them short-distance QA pairs (SQA); 
otherwise, long-distance QA pairs (LQA). 
It is obvious that matching LQAs is more difficult than SQAs. 
We assume that a two-party multi-turn dialogue contains two types of turns, 
questions (Q) and non-questions (NQ), 
which are labeled in advance~\footnote{We implemented a simple LSTM-based Q/NQ classifier
with accuracy equaling 96.10\% and F1-macro equaling 95.07\% on this dataset
which will be released to the research community.
Question detection is not the focus of this paper.}. 
Our task is to identify all answers from the set of NQs to a given Q. 

Previous methods~\cite{ding2008using,du2017discovering,jiang2018learning} on the task suffer from a major weaknesses: while classifying a pair of turns, they ignore the context of the turns in the dialogue. Meanwhile, their pre-defined features such as question words and answer words, are already implied by the Q and NQ labels in our definition and hence are not suitable for our task. He et al.~\cite{he2019learning} improves the above methods with a recurrent pointer network (RPN) model which takes the whole dialogue as an input. Their model was evaluated on a close-source customer service dialogue dataset. Although their model makes use of the context, they treat every utterance in the context equally with RNN-based networks which fails to capture the influences between turns especially with long distance. Besides, it encodes the distance information implicitly which downplays the effect of distance between the utterances. According to our experiments, none of the above approaches perform well on LQA pairs.

In this paper, we bring the dialogue context into the above simple models. 
For a given pair of Q and NQ to be matched, the context is defined as {\em history}, 
refering to the utterances between the Q and NQ. The critical part of our model 
is two simultaneous attention mechanisms that combine the history in a mutual way. 
Existing dialogue datasets, such as the Ubuntu dataset, do not contain
the QA matching labels. Moreover, they are either not two-party dialogues, or do not
contain long distance QA pairs. Therefore, we develop a new dataset based on
a Chinese online medical forum. We conducted experiments both on this dataset and the part of
labeled Ubuntu dataset~\footnote{The datasets and codes are available at: \url{https://github.com/JiaQiSJTU/QAmatching}}.

Our main contributions are as follows:
\begin{itemize}

    \item We aggregate dialogue history and distance information into a new deep
neural QA matching model. We show that distance is an important feature when encoded explicitly, and that utterances between Q and A can be effectively captured by a mutual attention mechanism (\secref{sec:method}). 
    \item Since there is no open source dialogue datasets designed for
QA matching task, especially for LQAs, 
we construct a reasonably sized dataset and release it to the research community (\secref{sec:data}). 
    \item The experimental results show that our proposed method outperforms 
other strong baselines, especially on LQAs. 
The techniques developed here are generic and can
be applied to other types of online dialogues (\secref{sec:results}).

\end{itemize}

\section{Problem Definition}
\label{sec:problem}

Our work aims at identifying the response turns to a question turn in a multi-turn,
two-party dialogue. Given a dialogue sequence with $T$ turns: 
$$[(R_1,L_1,U_1),(R_2,L_2,U_2),...,(R_T,L_T,U_T)]$$ 
where $R$ denotes the role, identifying which party utters
the turn. $L\in\{Q, NQ\}$, and $U$ is a sequence of words in natural language.

Our job is to match each $(Q, U_i)$ with corresponding $(NQ, U_j)$, where:

\begin{equation}
\begin{aligned}
j>i&\quad 1\leq i,j\leq T\\
R_i&\not=R_j\\
\end{aligned}
\end{equation}

The {\em distance} of a Q-NQ pair ($U_i$, $U_j$) is $j-i$. We define
the {\em history} as the turns 
$\{U_{i+1},U_{i+2}...,U_{j-1}\}$ which are located between the Q and NQ. The intuition will be explained in Section \ref{sec:mutual attention}.

Recent work by He et al.~\cite{he2019learning} considers a slightly different 
QA alignment problem where one answer can be matched with multiple questions. 
However, in this paper, we assume that if a question is asked repeatedly, the answer should be matched to 
the nearest question and all earlier ones are disregarded. 
In our definition, a Q can match nothing (U9) or several NQs (U2). From the viewpoint of a NQ, it is 
either matched, or not matched, with a Q (such as U7). When a NQ is matched to a Q, it is 
considered as an {\em answer} (A). Otherwise, it's considered as others(O).

\section{Approach}
\label{sec:method}

\begin{figure*}
	\centering
	\includegraphics[scale=0.40]{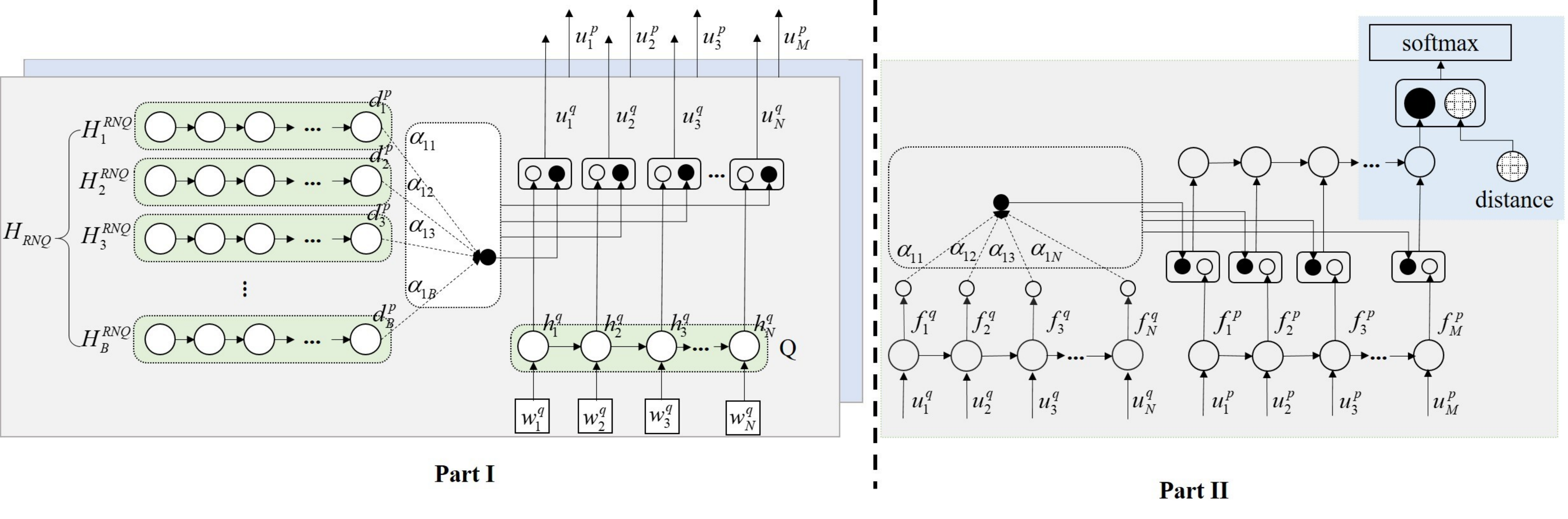}
	\caption{The architecture of the proposed match-LSTM based model with mutual attention mechanisms. \textbf{Part I:} Sentence Encoder and Mutual Attention.
The green part is Sentence Encoder and the rest is Mutual Attention. Front plate
(gray) shows the encoding of Q, while the back plate (blue) includes the encoding of NQ. \textbf{Part II:} Match-LSTM and Prediction Layer. The gray part is Match-LSTM and the blue part is Prediction Layer.}
	
	\label{fig:model1}
\end{figure*}

We propose an attention-based neural network model for QA matching 
in multi-turn dialogues between two parties. 
The model consists of four components (shown in Figure \ref{fig:model1}): 
\textit{Sentence Encoder} transforms the natural language turns into 
sentence-level embeddings. \textit{Mutual Attention} combines history turns 
based on two simultaneous attention mechanisms. 
\textit{Match-LSTM} is used to compare the processed sentence pair word by word. 
\textit{Prediction Layer} incorporates the distance information 
and calculates the matching probability. After calculating the probability for 
all Q-NQ pairs in a dialogue, a \textit{greedy matching algorithm} is employed to
match each NQ to zero or one Q.

\subsection{Sentence Encoder}
\label{sec:sentence-encoder}

Given an input turn as a sequence of words in natural language, we generate a neural representation using an LSTM \cite{gers1999learning}. The sentence encoder consists of pretrained word embeddings which are fed into a single LSTM layer. The output of all the hidden states or the last hidden state can both be regarded as the sentence-level embedding for this turn.

With the sentence encoder, we can get the neural representations for a Q-NQ pair and its history as follows:
\begin{equation}
\begin{aligned}
Q&=\{h_i^q\}_{i=1}^{N}\\
NQ&= \{h_i^p\}_{i=1}^{M}\\
H_{RQ}&=\{d_t^q\}_{t=1}^{A}\\
H_{RNQ}&=\{d_t^p\}_{t=1}^{B}\\
\end{aligned}
\end{equation}
where the number of words in $Q$ and $NQ$ are $N$ and $M$ respectively. $h_i$ represents the hidden state of each word. $H_{RQ}$ and $H_{RNQ}$ represent the history turns with the same role label as $Q$ and $NQ$ respectively. $A$ and $B$ are the number of turns in $H_{RQ}$ and $H_{RNQ}$, and $d_t$ is the last hidden state of each turn. Superscripts $p$ and $q$ are used to distinguish $Q$ and $NQ$. Here, we divide the history turns into 
two parts to support the idea of mutual attention between Q and NQ
in the following subsection.

The intuition for using different granularity of sentence embeddings is that we hope to 
keep more information for more important turns. Therefore, in order to calculate 
the matching score between a Q-NQ pair, we preserve all the hidden states for Q and NQ. 
The last hidden state of each turn in history is used to provide auxiliary information.

\subsection{Mutual Attention}
\label{sec:mutual attention}
To improve the prediction for each Q-NQ pair, 
naturally we take advantage of the dialogue history, 
especially the turns between Q and NQ. 
This idea comes from two considerations: i) if Q has been partially answered 
by another NQ in the history, then we should further explore 
whether the current NQ is a supplementary answer; ii) if there exists another Q which is closer 
to the NQ in both distance and semantics, the probability of matching 
the current Q-NQ pair should reduce. In a word, 
if the model can capture these intuitions, it's more likely to match 
the LQAs such as \{U2,U10\} without losing the accuracy on SQAs.

Besides, it should be noted that the question and matched answers should
definitely be uttered by different parties. 
In other words, the QA relations in a dialogue is focusing 
on the process of narrowing down the information gap between two parties, 
where the information interaction between parties is critical. 
So, we further divide the history into two parts: 
$H_{RQ}$ and $H_{RNQ}$ by different role labels as mentioned above. 
$Q$ is expected to interact with $H_{RNQ}$ while $NQ$ is expected to interact with $H_{RQ}$.

Borrowing the idea from Wang et al. \cite{wang2017gated}, we use two attention mechanisms to incorporate the history information into Q and NQ individually in a unified manner. For example, when dealing with the Q-NQ pair $\{U2,U10\}$, $H_{RQ}=\{U3,U7,U9\}$ and $H_{RNQ}=\{U4,U5,U6,U8\}$. The neural representation of $U2$ attends to $H_{RNQ}$ and $U10$ attends to $H_{RQ}$ simultaneously.
In other words, Q and NQ ``mutually'' attend to each other's history. Mathematically, as for $Q$ and $H_{RNQ}$: the $Q$ containing historical information can be obtained  via soft alignment of words in the question $Q=[h^q_1,h^q_2,...,h^q_N]$ and history turns $H_{RNQ}=\{d^p_1,d^p_2,...,d^p_B\}$ as follows (see Part I in Figure \ref{fig:model1}):

\begin{equation}
u^q_i=[h^q_i,c^q_i]
\end{equation}
where $c^q_i=att(h^q_i,H_{RNQ})$ is an attention-pooling vector of the whole history($H_{RNQ}$),
and $v$ and $W$ are the weights:
\begin{equation}
\begin{aligned}
s^i_j&=v^Ttanh(W_Qh^q_i+W_Hd^p_j)\\
a^i_k&=exp(s^i_k)/\sum_{j=1}^Bexp(s^i_j)\\
c^q_i&=\sum_{k=1}^Ba^i_kd^p_k
\end{aligned}
\end{equation}
Each word representation in $Q$ dynamically incorporates aggregated matching information from the history $H_{RNQ}$. 

The final representations of the question and the non-question after
mutual attention are $Q^\prime=[u^q_1,u^q_2,...,u^q_N]$ and 
$NQ^\prime=[u^p_1,u^p_2,...,u^p_M]$. Each vector
in $Q^\prime$ and $NQ^\prime$ not only represents the original turn meaning 
but also contains dialogue context. The effects of 
different choices of turns in history and the ways of aggregating the 
history will be discussed later.

\subsection{Match-LSTM}
We follow the work of Wang and Jiang~\cite{wang2016learning} 
and adopt match-LSTM to capture the features between the 
processed $Q^\prime$ and $NQ^\prime$ word by word.

As Part II of \figref{fig:model1} shows,
a one-layer LSTM is used to better fuse the history information at different steps.
We thus obtain $Q^{\prime\prime}=\{f^q_i\}_{i=1}^{N}$ and 
$NQ^{\prime\prime}=\{f^p_i\}_{i=1}^{M}$. 
We introduce a series of attention-weighted 
combinations of the hidden states of the question, where each combination is for a particular word in the non-question. The sentence-pair representation $P=\{p_i\}_{i=1}^{M}$ is calculated with attention mechanism as follows. 

\begin{equation}
    p_i=LSTM(p_{i-1},[f^p_i,c_i])
\end{equation}
where $c_i=att(Q,f^p_i,p_{i-1})$ is an attention-pooling vector of the whole question($Q$):

\begin{equation}
\begin{aligned}
s^i_j&=v^Ttanh(W_{NQ}f^p_i+W_Qf^q_j+W_pp_{i-1})\\
a^i_k&=exp(s^i_k)/\sum_{j=1}^Nexp(s^i_j)\\
c_i&=\sum_{k=1}^Na^i_kf^q_k
\end{aligned}
\end{equation}

Finally, we use $p_M$ to represent the whole Q-NQ pair which is used for predicting the final matching score.

\subsection{Prediction Layer }
At the last step, we use a fully-connected (FC) layer with softmax to do binary classification, which indicates whether this pair of turns has QA relation.

Driven by the intuition that distance is a critical feature 
for matching QA pairs, 
we explicitly add the distance at the end of our model to 
preserve its effectiveness. The distance $d$ is defined as 
a 10-dimensional one-hot vector, where the position of 
1 is equal to the distance. For example, if the distance is 4, then
the 4th dimension of the vector is set to 1. If $d\geq 10$, then the 10th
dimension is set to 1.

Finally, the probability is calculated as follows:
\begin{equation}
\begin{aligned}
FC&=W[p_M,d]+b\\
P(Q,NQ)&=Softmax(FC)
\end{aligned}
\end{equation}

In sum, the input of our model is a Q and NQ with associated information (history and distance), and the output of model is the probability which indicates if the Q is the matched question of the NQ. Hence, the loss function is the cross entropy between the predicted probability $P(Q,NQ)$ and the ground truth. The model can be seen as a binary classification model and all the parameters are trained altogether except the pretrained word embeddings.

\subsection{Greedy QA Matching}

Based on the trained model, every NQ now has a matching probability with every Q 
before it in the dialogue sequence. The greedy algorithm matches the NQ with the
Q with maximum probability if that probability exceeds 0.5.
The threshold is set to 0.5 because our model is actually a two-class classifier.

\section{Dataset Construction}
\label{sec:data}
Previous QA datasets are in the form of independent QA pairs~\cite{yang2015wikiqa} and do not
provide surrounding dialogue context. Although He et al.\cite{he2019learning} solved a similar problem, their customer
service dataset is not open to public due to privacy 
concerns. Wei et al.~\cite{wei2018task} 
published their dialogue dataset collected from an online forum. 
However, their work focuses on the dialogue policy learning and 
the data doesn't preserve the original utterances. 

There are also no published qualified dialogue datasets for the QA matching task. The IRC dataset~\cite{elsner2008you} and Reddit dataset~\cite{jiang2018learning} are both multi-party dialogues instead of two-party dialogues. Twitter Triple Corpus~\cite{sordoni2015neural} and Sina Weibo~\cite{shang2015neural} are not multi-turn dialogues 
with only two or three turns. 
MultiWOZ 2.0~\cite{budzianowski2018multiwoz}, CamRest676~\cite{wen2016network}, 
and Stanford Dialog Dataset~\cite{eric2017key} are multi-turn dialogues with dialogue act annotations but 
these QA pairs appear next to each other. 
If we shuffle the well-ordered dataset randomly or by some rules, it's unnatural and incorrect because the original dialogue context is destroyed as shown in Table \ref{tab:example}. 
The two-party Ubuntu dataset~\cite{lowe2015ubuntu} meets these requirements
and we annotated 1000 dialogue sessions. 
However, the statistics and experiments show that the Ubuntu dataset is not a good 
QA matching dataset especially for long distance QAs, given the limited manpower for
annotation. More details will be shown in \secref{sec:results}.

\begin{table}
	\scriptsize
	\centering
	\begin{tabular}{lcccc}
		\toprule[1.2pt]
		Turn & QA & Original & Shuffled\\
		\midrule[1pt]
		\tabincell{l}{r1:Where can I enjoy my holiday? I want to go \\ 
~~~~~somewhere near the sea and warm.}& Q1 & 1 & 1 \\

		r2:Maybe \textit{Xiamen} is a good choice.&   A1   &    2  &  4  \\

		 r1:Is \textit{there} anything delicious?
		 & Q2 & 3 & 2 \\
		r2:Wow, that’s quite a lot. There are……
		 & A2 & 4 & 3 \\
		\bottomrule[1.2pt]
	\end{tabular}
	\vspace{-0.25cm}
	\caption{An example of shuffling well-ordered dialogues. Unreasonable co-reference appears after shuffling.}
	\label{tab:example}
\end{table}

Hence, we create a new dataset suitable for this task. 
Nearly 160,000 distinct dialogues are collected from 
an online health forum\footnote{An dialogue example online: 
	\url{https://www.120ask.com/shilu/0cjdemaflhj6oyw9.html}}.
All the personal information was removed in advance by the website. 
After some basic data cleaning methods, 
we labeled 1000 randomly selected two-party multi-turn dialogues with 
Q (question), A (answer) and O (others) labels. 
A small amount of turns (0.24\%) are considered by the annotators 
to be both a question and an answer, and these are treated as questions
uniformly. Three graduated Chinese native students annotated the data with access to the Internet for searching in-domain knowledge. The Fleiss Kappa between three annotators 
was 0.75, indicating substantial agreement.

On average, each dialogue has 19.68 doctor turns and 17.32 patient turns. 
Most turns are made up of a sentence fragment, so the number of words 
for each turn is on average less than 10 
words~\footnote{There are on average 9.80 questions with 8.89 words, 
10.78 answers with 6.62 words, 16.41 casual chit chats with 6.99 words according to annotated dialogues.}. 
$21.9\%$ of the questions have no answers, $22.7\%$ of the questions 
have more than one answer and the remaining questions have the only answer. 
For questions that do have answers, each of them is matched to 1.41 answers 
on average.

The annotated dialogues are split into train/development/test sets by 7:1:2. 
The distribution of the QA pairs by distance is shown in Table \ref{tab:dataInfo}.

\begin{table}

    \centering
    \begin{tabular}{cccccc}
    \toprule[1.2pt]
    \diagbox{Dataset}{Distance} & 1 & 2 & 3 & 4 & $\geq5$ \\
    \midrule[1pt]
    Train  & 3439 & 2068 & 1029 & 450 & 554\\

    Development & 454   &   331   &    167  &  76   &  99  \\

    Test  & 947 & 592 & 274 & 136 & 168 \\
    \bottomrule[1.2pt]
    \end{tabular}
	\vspace{-0.25cm}
    \caption{The distribution of QA pairs by Q-A distances.}
    \label{tab:dataInfo}
\end{table}

We reconstructed the labeled dialogues into Q-NQ pairs with distance, history and binary golden label used for our models. A NQ from a party is paired with every earlier Q from the other party. If the pair is a QA pair, it is labeled as True(T). Otherwise, it is labeled as False(F). The distribution of positive and negative data on three datasets is shown in Table \ref{tab:pairdata}.

\begin{table}

        \centering
        \begin{tabular}{cccc}
                \toprule[1.1pt]
                \diagbox{Label}{Dataset} &Train& Development& Test\\
                \midrule[0.8pt]
                True &7540 & 1226  & 2116\\
  
                False & 80631 & 14889 & 23893  \\
                \bottomrule[1.1pt]
        \end{tabular}
    	\vspace{-0.25cm}
        \caption{The distribution of positive and negative Q-NQ pairs on three datasets.}
        \label{tab:pairdata}
\end{table}

\section{Experiment Setup}
\label{sec:eval}

In this section, we first list the baselines and the ablations of our full model. Next, we define the evaluation metrics. Finally, we show the details of hyperparameters in our model.

\subsection{Baselines and Our Method}

\subsubsection{Baselines} 
We mainly have the following four types of baselines.
\begin{itemize}
\item\textbf{Greedy strategy (GD)}
A simple baseline \textit{Greedy} is that, when a question is posed by $RQ$, we can directly match the following several NQs said by $\overline{RQ}$ as the answers. It stops when meeting another Q or a turn said by $RQ$. 
There are a few variants in this methods. GD1 selects only one satisfying answer, 
and GDN selects multiple satisfying answers. The methods with \textit{Jump} 
(\textbf{J}) skip the non-question turns uttered by $RQ$ when matching the 
NQs.

\item\textbf{Distance}
A simple model takes a 10-dimensional one-hot distance vector as the input of a fully-connected layer and outputs the score for each Q-NQ pair.

\item\textbf{Word-by-word match LSTM (mLSTM)}
This model is proposed by Wang et al.~\cite{wang2016learning}, used for natural language inference. It performs word-by-word matching based on an attention mechanism, with the aim of predicting the relation between two sentences.

\item \textbf{Recurrent Pointer Network (RPN)}
The model proposed by He et al.~\cite{he2019learning} is the previous
state-of-the-art method for a similar task, and is implemented with some modifications to fit our task. We use two parallel RPN to distinguish questions from two parties. Comparing the classification loss and regression loss proposed in this paper, we choose the one that performs best on our test set.

\end{itemize}

\subsubsection{Our Models}
By disabling some of its components, our model comes in three main
variants:
\begin{itemize}

    \item \textbf{Distance Model (DIS)} removes the mutual attention. It directly puts Q and NQ with the distance into Part II of the full model in Figure \ref{fig:model1}.
    \item \textbf{History Model (HTY)} disables the distance information at the prediction layer.
    \item \textbf{History-Distance Model (HDM)} is the full model we have explained in Section \ref{sec:method}.
\end{itemize}

\subsection{Evaluation Metrics}

Once we have identified all of the QA pairs, we count the true positive, 
false positive and false negative for each question. 
To measure the quality of the matched QA pairs, micro-averaging 
precision (\textbf{P}), recall (\textbf{R}) and F1-score (\textbf{F1}) are 
calculated,  with all questions in test dataset treated equally. 
Another metric, accuracy (\textbf{Acc}), is used when evaluating QA pairs matched
with a specific distance. Accuracy is calculated by the percentage of ground
truth QA pairs that are predicated positive by the models.

\subsection{Implementation Details}
 
We use Jieba\footnote{\url{https://github.com/fxsjy/jieba}} to do Chinese word segmentation and pre-train the 100 dimentional word embeddings with Skip-gram model~\cite{mikolov2013efficient} on all of the dialogues (including the unlabeled ones) we have collected online. For our proposed models, we use LSTM with hidden state size of 128 and 256 for Part I and Part II of the model respectively. We adopt Adam optimizer with 0.001 learning rate and 0.3 dropout. 
Learning rate decay is 0.95 and the training process terminates if the 
loss stops reducing for 3 epochs. All experimental results are 
averaged over three runs.

\section{Results and Analysis}
\label{sec:results}
 
In this section, we show the end-to-end results and ablation tests for the specific architectural decisions.
\subsection{Overall Performance}
The main results of different models are shown in Table \ref{tab:mainResults}. The last row lists the human performance, regarded as the upper bound of this task. Note that
human performance is not perfect due to inherent ambiguities in dialogues.

\begin{table}

	\centering
	\begin{tabular}{p{1cm}<{\centering}p{1cm}<{\centering}ccc}
		\toprule[1.5pt]
		Models &P&R& F1\\
		\midrule[1pt]
		GD1&69.84&44.73&54.53\\
		GDN  &70.03&69.11&69.57\\
		GD1+J&70.38&50.40&58.74\\
		GDN+J&51.47&82.90&63.51\\
		\hline
		mLSTM&58.17&4.20&7.84\\
		Distance&71.57&69.34&70.44\\
		RPN&72.40&68.63&70.46\\
\hline
		DIS&78.46$^\star$&70.34&74.70$^\star$\\
		HTY&75.40$^\star$&76.42$^\star$&75.90$^\star$\\
		HDM&76.44$^\star$&78.44$^\star$&\textbf{77.43}$^\star$\\
		\hline
		Human &85.11&84.21&84.66\\
		\bottomrule[1.5pt]
	\end{tabular}
	\vspace{-0.25cm}
	\caption{The end-to-end performance of all methods on test dataset. 
Scores marked with $^\star$ are statistically significantly better than the RPN 
with $p<0.01$.} 
	\label{tab:mainResults}
\end{table}

The results of the rule-based methods are not bad, which indicates that questions 
are actually followed by their answers in many cases. The GDN increases the F1-score 
to 69.57\% compared with Greedy-1 because it can solve the case of simple fragmented 
answers. For GDN+J, the recall is the best among all the methods while precision and 
F1-score suffer. 
The reason is that GDN tends to match NQ with Q as much as possible, so many chit chats will be regarded as answers, which reduces the precision.

Model mLSTM underperforms because it is difficult to solve the QA matching problem with 
only two short texts without history. The word distribution between the questions 
and answers are quite different and maybe unrelated without background knowledge. 
Distance achieves good scores which shows that the distance is a very important factor 
when identifying QA relations in dialogues. People tend to answer a question 
the moment they see it except in the case of incremental QAs. RPN achieves competitive 
results. It mainly benefits from taking the dialogue session as a whole which 
contains all the information in a session. 
 
Our proposed models achieve the best results compared with above models. 
The HDM improves the F1-score to 77.43\%, significantly better than RPN by t-test with
$p<0.01$. 
Although the recall of HDM is not better than GDN+J and the precision is 
lower than DIS, the overall quantity and quality of QA pairs identified 
are the best, shown by the highest F1-score. 
In addition, the comparable results achieved by HTY demonstrate that 
QA matching not only depends on 
the distance but also relies on the history information.
This shows that HDM model successfully combines both the distance and history 
information.

\subsection{Variable Distance Matching}

 According to the results above, we can find that the model Distance, RPN, DIS, HTY and HDM are competitive. Thus, we further analyze the accuracy of these five models on variable distances.

\begin{table}

	\centering
	\begin{tabular}{p{1.5cm}<{\centering}ccccc}
		\toprule[1.3pt]
		 Models &1&2&3&4&$\geq5$\\
		\midrule[1pt]
		Distance      &\textbf{100.0}&88.01&0.0&0.0&0.0 \\
		RPN  &89.37&69.37&50.12&36.96&13.10\\	
		DIS &96.23&\textbf{89.13}$^\star$&17.03&2.45&0.0\\
		HTY &94.37&78.89$^\star$&57.42&38.48&\textbf{28.17}$^\star$\\
		HDM &95.99&83.16$^\star$&$\textbf{59.37}^\star$&\textbf{40.44}&24.80$^\star$\\
		\bottomrule[1.3pt]
	\end{tabular}
	\vspace{-0.25cm}
	\caption{The Acc (\%) of matched QA pairs on variable distances.}
	\label{tab:longrangeResults}
\end{table}
Table \ref{tab:longrangeResults} shows that Distance and DIS work well on SQAs 
but deteriorate rapidly 
as the distance grows, indicating that relying solely on 
the distance information is insufficient. 
On the other hand, using dialog context, 
the matching accuracy of RPN and HTY is generally lower on SQAs but higher on 
LQA. Our full model (HDM) is actually a good trade-off 
by incorporating both distance and history information. This also accords 
with the decision process made by human annotators.

We also conduct a Z-test on the 
results to show that the improvements on the LQAs are statistically 
significant even with a small sample size.

\subsection{Ablation Tests}
We justify the design of our model in the following two aspects.

\subsubsection{Different definitions of history}

To show the effectiveness of using the turns between Q and NQ 
as history, we devise variants on the HDM model 
for comparison:
\begin{itemize}
	\item \textbf{Q-history Model (QH)} has the same structure as HDM where the history is all the turns before Q.
	\item \textbf{A-history Model (AH)} has the same structure as HDM where the history is all the turns before NQ. 

\end{itemize}

The main results with different choices of history are shown in 
Table \ref{tab:historychoice}.
Our final model (HDM) outperforms QH and AH, indicating that the turns between Q and NQ are significant when figuring out the relation of a Q-NQ pair. The turns before Q is actually not that important for matching Q and NQ. 
Although there is an overlap between the history we defined and 
the turns before NQ, 
the turns before Q brings more noises than benefits for
the end-to-end performance. This suggests that our definition of 
history as the turns between Q and NQ is reasonable and effective.

\begin{table}
	
	\centering
	\begin{tabular}{p{1.5cm}<{\centering}|c|ccc}
		\toprule[1.3pt]
		Models &F1&Acc@3&Acc@4&Acc@$\geq5$\\
		\midrule[1pt]
		QH&74.56&21.53&10.61&0.79 \\	
		AH&73.84&15.81&10.78&4.76 \\
		HDM&\textbf{77.43}&\textbf{59.37}&\textbf{40.44}&\textbf{24.80}\\
		\bottomrule[1.3pt]
	\end{tabular}
	\vspace{-0.25cm}
	\caption{The matching results on different choices of history.}
	\label{tab:historychoice}
\end{table}

\subsubsection{Different ways of attending to the history}

To evaluate the effectiveness of mutual attention for aggregating the 
history, we devise variants of the HDM model for comparison:
\begin{itemize}
	\item \textbf{Non-mutual Model (NM)} has the same structure of HDM 
	where $Q$ attends to $H_{RQ}$ and $NQ$ attend to $H_{RNQ}$.
	\item \textbf{Identical history model (ID)} has the same structure of HDM where $Q$ and $NQ$ attend to the same history $H_{RQ}\bigcup H_{RNQ}$.

\end{itemize}

The main result in Table \ref{tab:historyways} reveals that our choice 
of separating the history by role label and mutually attending to each other
does work. The full model (HDM) consistently outperforms both NM and ID.

For the ablation test on different ways of attending to the history, we conclude that
NM is better than ID. It's due to better understanding on individual speakers which can
help the understanding of the dialogue, similar to the idea in Shi and Huang's work~\cite{ShiH19}.
Our work targets the QA relations in dialogues which are more related to the interactions 
between speakers. As a result, our full model is better than NM. 

The difference between our full model HDM and ID is that in ID both Q and NQ attend to
all turns in the history regardless of who uttered those turns, whereas HDM employs a
mutual attention mechanism that distinguishes turns by their speakers. Specifically, the
Q only attends to those turns uttered by the NQ speaker, while the NQ attends to those
turns by the Q speaker. This resembles to some extent the firm attention in Amplayo's paper\cite{amplayo2018entity}.
This will help the model to focus on the interactions between speakers.

\begin{table}

	\centering
	\begin{tabular}{p{1.5cm}<{\centering}|c|ccc}
	\toprule[1.3pt]
	Models &F1&Acc@3&Acc@4&Acc@$\geq5$\\
	\midrule[1pt]
	NM&75.81&57.18&37.50&20.04 \\	
	ID&75.46&54.50&28.43&11.70\\
	HDM&\textbf{77.43}&\textbf{59.37}&\textbf{40.44}&\textbf{24.80}\\
	\bottomrule[1.3pt]
\end{tabular}
	\vspace{-0.25cm}
	\caption{The results on different ways of aggregating history.}
	\label{tab:historyways}
\end{table}

\subsubsection{Example Outputs}

\begin{CJK}{UTF8}{gbsn}
To provide a better understanding of the behavior of our models, we include an example 
output in Table \ref{tab:case1}. It contains both LQAs and SQAs. 
In this case both HYD and HDM predict QA relations better than the baseline RPN. 
As for SQAs, all of the models perform well. However, the DIS model is 
obviously not capable of matching LQA pairs. It indicates that the distance information sometimes hurts the performance of HDM on matching LQA pairs.

	\begin{table*}

		\small
		\centering
		\begin{tabular}{p{1.5cm}<{\centering}cccccc}
			\toprule[1.3pt]
			Ground Truth &RPN&DIS&HTY&HDM&Role&Utterances\\
			\midrule[1.3pt]
			\multicolumn{5}{c}{Q1}&P&\makecell{Boy, 4 months. He tried a little yolk yesterday and shat that night \\but haven't shat until today what's wrong????}\\
			\hline
			O &O &O &O &O &D&\makecell{Hello}\\
			\hline
			O &O &O &O &O &P&\makecell{Hello}\\
			\hline
			\multicolumn{5}{c}{Q2}&D&\makecell{Is he four months old}\\
			\hline
			A2&A2&A2&A2&A2&P&\makecell{Yes}\\
			\hline
			A1&A1&O &A1&A1&D&\makecell{Eat too early}\\
			\hline
			A1&O&O &A1&A1&D&\makecell{Not advise}\\
			\hline
			A1&O&O &A1&A1&D&\makecell{Difficult for digestion}\\		
			\bottomrule[1.3pt]
		\end{tabular}
		\caption{A correct case of predictions and human annotations in our dataset.}
		\label{tab:case1}

	\end{table*}
	
\end{CJK}

\subsection{Results on the Ubuntu Dialogue Corpus}
\label{sec:uresults}

The Ubuntu dataset~\cite{lowe2015ubuntu} is a large unannotated dialogue corpus designed for next utterance classification and dialogue generation. It is extracted from multi-party dialogues, while conversation disentanglement is still an unsolved problem. 
Therefore, many dialogues in this dataset don't make much 
sense because the two parties are not actually talking to each other (they were
talking to a third party whose utterances were removed). 
It maybe a good dialogue resource with 930,000 dialogues, but it is 
impossible to label such large amount of dialogues.

We randomly sampled 1000 dialogues each with more than 10 turns. 
Then we annotated the extracted corpus with (Q, A, O) labels as mentioned above. 
If the current dialogue contains no Q-NQ pairs, it is replaced by a new dialogue 
with similar number of turns. Finally, we collected 1000 annotated dialogues. 
The QAs in Ubuntu corpus mainly focus on step-by-step operations and the dialogues are 
lack of LQAs. The results are showed in Table \ref{tab:mainResults-ubuntu}.

\begin{table}

	\centering
	\begin{tabular}{p{1cm}<{\centering}p{1cm}<{\centering}ccc}
		\toprule[1.5pt]
		Models &P&R& F1\\
		\midrule[1pt]
		GD1&89.33&58.96&71.03\\
		GDN  &84.62&78.86&81.64\\
		GD1+J&83.62&65.02&73.16\\
		GDN+J&57.05&89.49&69.68\\
		\hline
		mLSTM&64.88&0.37&0.73\\
		Distance&84.97&78.99&81.87\\
		RPN&71.33&61.80&66.23\\
		\hline
		DIS&85.36&66.21&74.57\\
		HTY&85.25&77.53&81.20\\
		HDM&85.88&77.53&81.48\\
		\hline
		Human &90.28&84.42&87.25\\
		\bottomrule[1.5pt]
	\end{tabular}
	\vspace{-0.25cm}
	\caption{The end-to-end performance of all methods on the Ubuntu test dataset.}
	\label{tab:mainResults-ubuntu}
\end{table}

The results of Distance and GDN achieve the top-2 highest F1 score, 
which is consistency with the fact that 41.95\% of questions have no answers 
and 79.55\% of the QA pairs are consecutive in the dialogue. 
However, it still shows the effectiveness of our full model HDM, 
which achieves the competitive results to the best results.

\section{Related Work}
\label{sec:relatedwork}

Detection of QA pairs from online discussions has been widely researched these years. Shrestha and Mckeown~\cite{shrestha2004detection} learned rules using Ripper for detecting QA pairs in email conversations. Ding et al.~\cite{ding2008using}, Kim et al.~\cite{kim2010tagging} and Catherine et al.~\cite{catherine2012does} applied the supervised learning method including conditional random field and support vector machine. Cong et al.~\cite{cong2008finding} proposed an unsupervised method combining graph knowledge to solve the task. Catherine et al.~\cite{catherine2013semi} proposed semi-supervised approaches which require little training data. He et al.~\cite{he2019learning} used the pointer network to find QA pairs in Chinese customer service. 
However, the tasks mentioned above are all different from ours. 
We identify QA pairs from two-party dialogues on online discussion forum, 
and focus especially on long-distance QA pairs. Besides, our 
dialogue is constrained between two roles who can both utter questions and  answers.

There exists several methods on other tasks which can be adapted to 
our QA matching problem. Feature-based method is popular for solving 
many NLP problems. In the work of Ding et al.~\cite{ding2008using}, 
Wang et al.~\cite{wang2010modeling} and 
Du et al.~\cite{du2017discovering}, 
they examined lexical and semantic features in two sentences 
for QA matching. However, the features such as common question words 
and roles have already been explicitly annotated in our data. 
Besides, other features such as special word occurrence or time stamp are 
unavailable here. According to the data, we considered the distance as 
the most important feature and implemented this feature-based method as one 
baseline. Recent researches using deep neural networks have increased a lot. 
He and Lin~\cite{he2016pairwise} and 
Liu et al.~\cite{liu2016modelling} used the sentence pair interaction 
approach which takes word alignment and interactions between the sentence 
pair into account. Attention mechanism was also added for performance 
improvement~\cite{rocktaschel2015reasoning,wang2016learning,chen2017enhanced}. 
We also use word alignment and interactions to calculate the QA similarity. 
Specially, we adopt attention mechanism to solve the LQA cases.

There are other kinds of alignment problems such as temporal sequences alignment. Video-text alignment is one of the temporal assignment or sequence alignment problems. Previous work automatically provided a time (frame) stamp for every sentence to align the two modalities such as \cite{bojanowski2015weakly} and \cite{dogan2018neural}. Bojanowski et al.~\cite{bojanowski2015weakly} extended prior work by including the alignment of actions with verbs and aligned text with complex videos. Dynamic time warping (DTW) is anothor algorithm for measuring similarity between two temporal sequences. It's also widely used in video-text alignment task~\cite{dogan2018neural}, 
speech recognition task~\cite{vintsyuk1968speech}.

\section{Conclusion}
\label{sec:conclusion}
In this paper, we focus on identifying QA pairs in two-party multi-turn online dialogues based on turns with Q or NQ labels. Our proposed models achieve the best accuracy overall, and perform particularly well on LQAs. We also discuss the model decisions of using two attention mechanisms in a mutual way and the definition of history. Our future work will focus on more discourse relations in dialogues. Utilizing out-of-domain knowledge is another research direction for utterances matching task, especially for health-related dialogues.

\section*{Acknowledgement}
This research was supported by NSFC grant 91646205, Didi-SJTU Joint Research Scheme, and the computing facilities in the 
Network and Information Center of Shanghai Jiao Tong University.

\bibliography{ecai}
\end{document}